# LIVER SEGMENTATION IN CT IMAGES USING THREE DIMENSIONAL TO TWO DIMENSIONAL FULLY CONVOLUTIONAL NETWORK

*Shima Rafiei[1], Ebrahim Nasr-Esfahani[1], S.M.Reza Soroushmehr[2,3],*
*Nader Karimi[1], Shadrokh Samavi[1, 3], Kayvan Najarian[2,3]*

[1]Department of Electrical and Computer Engineering, Isfahan University of Technology, Isfahan 84156-83111, Iran
[2]Department of Computational Medicine and Bioinformatics, University of Michigan, Ann Arbor, MI, U.S.A.
[3]Michigan Center for Integrative Research in Critical Care, University of Michigan, Ann Arbor, MI, U.S.A.

**ABSTRACT**

The need for CT scan analysis is growing for pre-diagnosis and therapy of abdominal organs. Automatic organ segmentation of abdominal CT scan can help radiologists analyze the scans faster and segment organ images with fewer errors. However, existing methods are not efficient enough to perform the segmentation process for victims of accidents and emergencies situations. In this paper we propose an efficient liver segmentation with our 3D to 2D fully convolution network (3D-2D-FCN). The segmented mask is enhanced by means of conditional random field on the organ's border. Consequently, we segment a target liver in less than a minute with Dice score of 93.52.

*Index Terms*— Liver segmentation, deep learning, 3D-2D-FCN, conditional random field

## 1. INTRODUCTION

Liver, the largest abdominal organ, is at the risk of trauma or physical injuries. Since this organ is vital for human life, medical clinics have to be quick enough in detection of its internal trauma. For detection of any injury or bleeding in abdomen the best modality is computed tomography (CT) as it can reveal internal trauma in a painless and accurate way, that helps saving patients' lives. In these cases, radiologists have to analyze slice by slice of abdominal scans which might be a tedious job by increasing the number of referrals in a clinic center. Therefore, the automatic organ segmentation is a kind of beneficial computer-aided diagnosis and therapy which in emergency cases can help accelerate the process of trauma detection.

For this goal, a large number of approaches have been proposed in the literature for organ segmentation. Most of the organ segmentation methods are atlas-based. Probabilistic atlas[1] and multi-atlas label fusion [2][3] are examples of atlas-based methods. Probabilistic methods exploit the probability of location and shape prior maps of organs by considering all atlases. Multi atlas methods perform atlas selection for a specific target and perform weighted voting on atlas patches [3], entire atlas, organ, or voxels hierarchically [2]. These methods are capable of capturing useful information among atlases leading to robust segmentation. Nevertheless, atlases consist of different sizes and directions with various organ shapes, locations and appearances. Hence, for having useful atlases related to a target, they essentially need to be registered with the target usually in two steps of affine translation and non-rigid deformation to attain target conditions [4]. However, deformation step, applied in most of the atlas-based approaches, is very time consuming as a pre-processing step and employing such approaches in medical frameworks cannot be applicable especially in an emergency situation.

The state-of-the-art approaches omit the role of registration step for the organ segmentation, thanks to different deep learning methods. Roth *et al.* [5], apply sliding window-based networks to segment pancreas organ hierarchically. They use ConvNet networks in the form of coarse-to-fine to classify pancreas patches and superpixels. The initial set of superpixels is obtained using a random forests-based approach. For pancreas segmentation, Heinrich and Oktay [6] propose a deep network named BRIEFnet with a binary sparse convolution as its first layer. This layer has low complexity with large receptive field on a 3D ROI of a scan, beneficial for 3D network with large parameters. A deep 3D-CNN is proposed in [7] to obtain a prior map for liver segmentation. In order to refine the under and over segmentation of the map, it incorporates local and global statistics from the prior segmentation and optimizes final results. Authors of [8] propose a deep supervised network for liver segmentation. The input to the network is a partial 3D bounding box which has to be slid on a target scan during the test time. To alleviate the issue of vanishing gradient, this network up-scale feature maps resulted from two middle layers by means of additional deconvolutional layers and calculate the gradient of loss from several branches. In this paper, we intend to reduce the processing time for the segmentation of 3D medical images. Most of the existing algorithms use 3D data pieces and carry out the volume through the network. We propose to convert the 3D pieces into 2D which facilitates the process of training and reduces the memory consumption. We apply our method for automatic liver segmentation and exploit some strategies to achieve more accurate segmentation mask. Then afterward by employing conditional random field (CRF) [9] locally on the organ boundaries, we enhanced the segmentation of liver organ with Dice of 93.52 and compare our method on

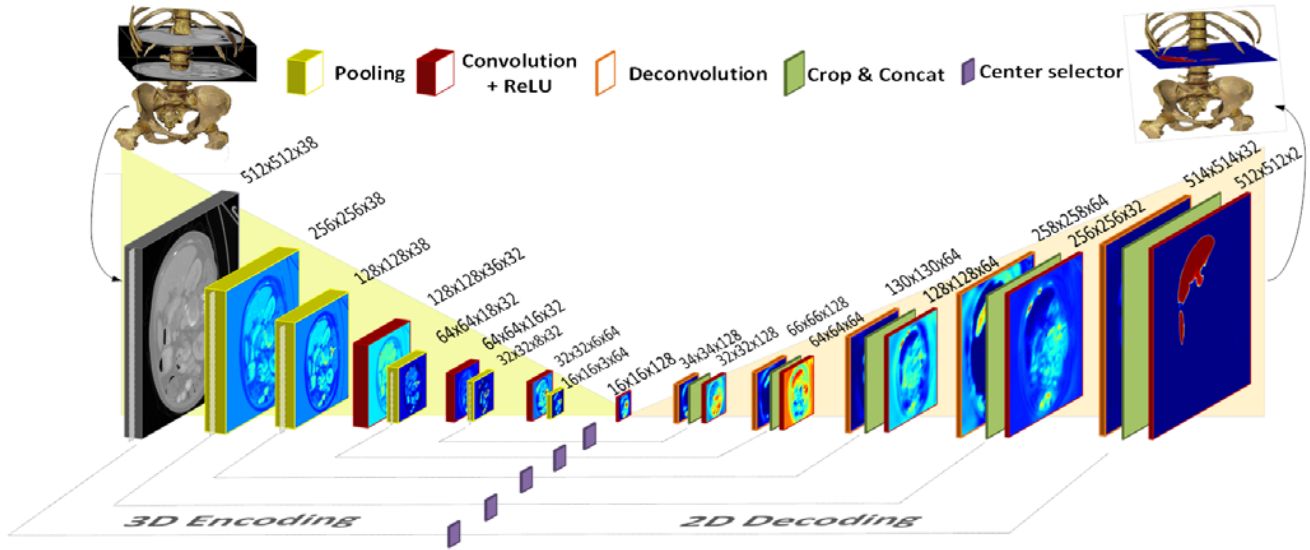

Fig. 1. Proposed FCN-based network for the liver segmentation. The channel number is annotated for each layer. No channel is shown to simplify illustration.

## 2. PROPOSED METHOD

the training set of MICCAI 2015 with the state-of-the-art method of [3]. This method can be executed in emergency medical centers with an efficient time i.e. less than a minute. The remaining of this paper is organized as follows. In section II, we describe our deep network architecture along with strategies effective for our network training. In section III, we express a graphical post-processing model used for enhancement of segmentation mask whereas in section IV we indicate the qualitative and quantitative results of our approach with comparison to a state-of-the-art method. In section V we conclude the paper.

The inter-patient variety of shapes and locations seems to be extremely large. In presence of large organ variety among patients, the organ detection would be challenging. With this regard whenever the detection phase of a target organ is not accurate enough, the segmentation algorithm would mislead the labels of an organ. To overcome this issue, most of the approaches tend to perform organ registration to achieve a relatively appropriate location prior map. However, this time consuming preparation step makes these methods inappropriate for medical tools. Another approach that can be used is to find locations of organs in terms of bounding boxes with decision forest [10] or regression forest [11] approaches, in a reasonable time. However, such methods are not robust enough to detect organs with large variations in shapes, locations and appearances despite their use of hand-craft features.

### 2.1. Network architecture

In this paper, the detection and segmentation of liver is directly performed by our designed 3D-2D-FCN with the architecture illustrated in Fig.1. As can be observed, a number of sequential slices of one scan are fed into the network and as a result the probability map corresponding to the middle slice is generated at the end of the network.

Overall, our network has two major parts. It starts with the 3D encoding phase and ends with a 2D decoding phase. In the encoding phase, by means of 3D kernels, a zoom-out 3D receptive field on the volume is created. It is shown that the receptive filed can better determine the boundaries and detect a solid large organ. In contrast to 2D kernels, 3D kernels are capable of capturing 3D surface of an organ, and perform more powerful organ detection and segmentation. In other words, applying 3D kernels with large receptive fields on sequential slices of a scan would generate discriminative feature maps which contain the 3D shape and surface of the related organ, rather than generating only 2D curves. This would also mitigate the issues of large variations among organ slices. In fact, detection of the first and last slices of liver, which do not contain shape information of this organ, is challenging. This challenge is due to the fact that such slices contain only a small part of the liver surrounded by other organs with vanishing borders. Hence, that small portion of the liver can be missed among neighboring viscera. In this case, our network comprehends the continuity of sequential slices and does not miss small malformed tissue of the liver.

At the end of the encoding phase, our network employs 2D kernels, rather than 3D kernels. This is done to reduce memory consumption by producing the segmentation mask only for the middle slice. In this situation, the network can determine the segmentation mask of a middle slice using the presence of a significant number of neighboring slices.

In each iteration, a part of a scan with the size of $512 \times 512 \times 38$ is fed into the network. The network pools the first two dimensions for decreasing the resolution, which leads to a larger field of view and reduces the memory consumption. Feature map dimension is reduced by passing through several stages of pooling layers. The last feature map, achieved at the end of the encoding layers, is $16 \times 16 \times 1$ which shows the loss of its third dimension. In addition to achieve the probability of each pixel in the middle slice, the last feature map, which is obtained by the encoding phase, has to be up-sampled to reach the same size as the ground truth. Hence, in the decoding phase, the network continues up-sampling by 2D convolutional and deconvolutional layers. In this way, for better reconstruction of missing features the incorporation of the information in former layers can be beneficial [12][13]. In the decoding phase, we up-sample feature maps, step-by-step with the help of deconvolution layers. Hence, each layer in the decoding phase corresponds to a layer in the encoding phase with almost the same size. However, the corresponding feature maps in the encoding phase are 3D while all feature maps in the deconvolutional layers are 2D. To solve this inconsistency, we select only the center slice of each channel. Afterward the center slices are fed into *crop&concat* layer, followed by convolutional layer for interfering new feature maps.

In this paper, all 3D and 2D convolutional layers use kernels with the size of 3×3×3 and 3×3 respectively with stride of one and zero-padding of size one. Thus the convolutional layers do not change their input size. Furthermore, for all deconvolutional layers we set the kernel size to $4 \times 4$ in order to up-sample feature maps while the stride size is set to 2 for better corporation of neighboring features for construction of missing ones. The strategy of the kernel size with smaller stride steps might be more beneficial for the precise reconstruction of deconvolutional layers.

## 2.2. Well-training strategies

Although our network is deep enough and is capable of learning proper parameters for liver segmentation, it is prone to over-fitting issues. To mitigate this issue and boosting the training phase, we employ some techniques effective for well training of deep networks.

In this regard, one of the most challenging issues is the presence of fuzzy borders among organs. For example, surrounding the liver, there are some organs such as right kidney, gallbladder, stomach, heart and pancreas. Following the boarders of these organs, in many cases, is even difficult for a human. Vanishing boundaries are due to similar appearance, color intensities, textures of these neighboring organs, and presence of noise and artifacts in CT images. These parts of scans are more prone to error in segmentation results. To address these issues, according to (1), we use cross entropy loss function multiplied by a weighing map. The role of the weighing map is to magnify the gradient of loss on the boundaries of an organ during the training phase.

$$E = -\sum_{x \in \mathbb{Z}^2} w(x) \log(p_c(x)) \qquad (1)$$

where, $w(x)$ is the weighting map that is calculated using (2). By means of (2) we calculate the weighting map of each slice, where $d(x)$ is the distance between the pixel $x$ and the organ boundaries. By using this equation, the closer a pixel is to the boundary, the higher loss is imposed to the network. On the other hand, the closer the pixel is to the border, the greater its importance would be in the learning process of the network.

$$w(x) = 1 + w_0 \cdot \exp\left(-\frac{d(x)}{2\sigma^2}\right) \qquad (2)$$

With regard to this strategy our network errors on boundaries would significantly decrease.

In addition, for increasing the speed of network, after each convolution layer we employ batch normalization technique [14]. Furthermore, in order to mitigate over-fitting issue, we use dropout technique, with probability of 0.5, right between the encoding and decoding phases of the network.

In an FCN-based network, the back-propagation errors are per voxel. In fact, each voxel of an input would be treated as a new sample for FCN. Thus the number of training sets fed into the network is equivalent with the overall voxels of the training set. Hence despite the lack of CT scans in this work, the issue of over-fitting has small effect in our network. We compensated small number of samples by augmenting each scan into seven scans. For a realistic augmentation, we rotate each scan from -30 to 30 degrees in steps of 10. This range of rotations is as if a patient, or even a part of the abdomen, leans to left or right during the scanning process. Larger than 30 degrees rotations would create synthetic scans which are not needed in the test phase and might mislead the network.

## 3. POST-PROCESSING

To enhance the segmentation mask on the boundary areas, we employ fully-connected CRF model [9] on pixels of the organ boundary regions in each slice. For this aim, we employ the energy function shown in (3) on a graph $X$ involving boundary pixels. This energy function contains a unary potential of pixel $i$ and pairwise potential defined on its neighbor region, $N_i$.

$$E(X) = \sum_i \psi_u(x_i) + \sum_{i,j \in N_i} \varphi_p(x_i, x_j) \qquad (3)$$

In (3), $x$ is a label assigned to each pixel, and $\psi_u(x_i)$ is the unary potential term equal to $-log\, P(x_i)$ and $P(x_i)$, produced by our 3D-2D-FCN, is the probability of pixel $i$ belonging to the liver. In (4), the pairwise term $\varphi_p(x_i, x_j)$ contains bilateral and unilateral kernels, while this term penalizes only the pixels with non-similar labels i.e. $\mu(x_i, x_j) = [x_i \neq x_j]$. In (4), $p$ indicates position and $I$ shows gray-scale value. The bilateral kernel motivates nearby pixels with similar intensity to be in the same class while unilateral is used for smoothness and elimination of isolated regions. Furthermore, $\theta_\alpha$, $\theta_\beta$ and $\theta_\gamma$ are hyper parameters that have been set for controlling the scale of kernels.

$$\varphi_p(x_i, x_j) = \mu(x_i, x_j) \left[ w^{(1)} \exp\left(-\frac{|p_i - p_j|^2}{2\theta_\alpha^2} - \frac{|I_i - I_j|^2}{2\theta_\beta^2}\right) + w^{(2)} \exp\left(-\frac{|p_i - p_j|^2}{2\theta_\gamma^2}\right) \right] \quad (4)$$

By applying fully-connected CRF on only the boundary areas with restricted neighborhood, the execution time is significantly reduced.

## 4. EXPERIMENTAL RESULTS

### 4.1. Dataset

We use training dataset of abdominal CT scans provided by MICCAI 2015 [15]. The number of scans in the training set is 30 scans along with corresponding ground truth maps. The whole scans consist of 3631 slices. This dataset has $(512 \times 512 \times 85) \sim (512 \times 512 \times 198)$ voxels captured during portal-venous contrast phase while patients were suffered from retrospective ventral hernia or ongoing colorectal cancer.

### 4.2. Implementation details

In this paper, we generate our liver segmentation on a training set of 30, using leave-one-out cross-validation strategy. Each time we consider one fold as a testing set and four folds for the training set through which two scans are chosen randomly as a validation set. In addition, we employ early stopping technique in the training phase in order to avoid the over-fitting issue. The proposed method is implemented on a system with an NVIDIA GeForce GTX Titan X GPU card, Intel Core i7- 4790K processor, 32 GB of RAM. In this paper, we set $w_0$ and $\sigma$ to 20 and 30 respectively. We also set parameters of CRF such as $w^{(1)}$, $w^{(2)}$, $\theta_\alpha$, $\theta_\beta$ and $\theta_\gamma$ to 2, 0.5, 0.01, 20, and 20 respectively. Moreover, we consider the $N_i$ neighbor region of CRF as 5×5 area. In this paper, we evaluate our method on CT scans of 10 first patients in Table1, to compare with the reported results of [3] provided by the author. Our method significantly reduced the test time with more accurate segmentation performance. In this comparison, the time has been reported for 100 slices with size of 512×512.

**Table 1 the quantitative comparison of liver segmentation on 10 CT scans of MICCAI challenge.**

| Segmentation performance | | |
|---|---|---|
| segmentation algorithms | Dice Score (%) | Time (second) |
| Heinrich[3] | 92.95 | 1101 |
| 3D-2D-FCN | 92.80 | **42.72** |
| 3D-2D-FCN + CRF | **93.52** | 55.59 |

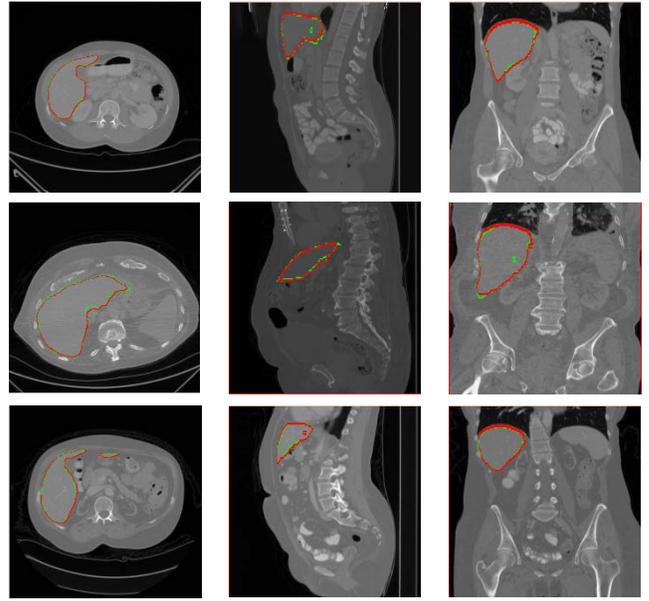

Fig. 2. Each row belongs to a scan with the axial, sagittal and coronal view. The final segmentation boundary achieved by this paper is shown in red while the ground truth boundary is in green.

## 5. CONCLUSION

In this paper, we proposed 3D-2D-FCN for automatic liver segmentation. The 3D encoding phase is for capturing 3D surfaces while 2D decoding phase reduces the memory consumption and facilitates the training process. Moreover, we employed effective strategies for well-training of the network and enhanced the network's results on the boundaries of the liver organ by using CRF. Our proposed method has small processing time and hence can be applicable in emergency clinical centers. Also, the proposed strategy can be generalized applied to other 3D medical imaging systems.

## 6. ACKNOWLEDGMENT